\documentclass[10pt,twocolumn,letterpaper]{article}

\usepackage{cvpr}              

\usepackage{graphicx}
\usepackage{amsmath}
\usepackage{amssymb}
\usepackage{booktabs}
\usepackage{algorithm, algorithmic}
\usepackage{multirow}
\usepackage{wasysym}
\usepackage{pifont}
\usepackage{threeparttable}
\usepackage{color}

\usepackage{tablefootnote}

\usepackage{sidecap}
\sidecaptionvpos{figure}{c}

\usepackage{paralist}

\definecolor{xf}{RGB}{69,137,148}
\definecolor{yellow2}{RGB}{235,213,52}
\newcommand{\xf}[1]{{\color{black} #1}}
\newcommand{\xff}[1]{{\color{black} #1}}

\usepackage[pagebackref,breaklinks,colorlinks]{hyperref}

\usepackage[capitalize]{cleveref}
\crefname{section}{Sec.}{Secs.}
\Crefname{section}{Section}{Sections}
\Crefname{table}{Table}{Tables}
\crefname{table}{Tab.}{Tabs.}

\def\cvprPaperID{9294}
\def\confName{CVPR}
\def\confYear{2023}

\newcommand{\net}{UnSniffer\xspace}
\newcommand{\MyScore}{Generalized Object Confidence Score\xspace}

\newcommand{\ood}{unknown objects\xspace}

\begin{document}

\title{Unknown Sniffer for Object Detection: Don't Turn a Blind Eye\\ to Unknown Objects}

\author{Wenteng Liang$^{1,\dag}$, Feng Xue$^{1,\dag}$, Yihao Liu$^1$, Guofeng Zhong$^2$, Anlong Ming$^{1,*}$\\
$^1$Beijing University of Posts and Telecommunications, China\\
$^2$Chongqing University of Posts and Telecommunications, China\\
{\tt\small \{liangwenteng,xuefeng,l1h,mal\}@bupt.edu.cn}}

\maketitle

\renewcommand{\thefootnote}{\fnsymbol{footnote}}
\footnotetext[2]{Equal Contribution}
\footnotetext[1]{Corresponding Author}

\footnotetext{This work was supported by the national key R \& D program intergovernmental international science and technology innovation cooperation project 2021YFE0101600.}

\begin{abstract}
The recently proposed open-world object and open-set detection \xf{have achieved} a breakthrough in finding never-seen-before objects and distinguishing them from \xf{known ones}.
However, their studies on knowledge transfer from known classes to unknown ones \xf{are not deep enough},
\xf{resulting in} the scanty capability for detecting unknowns hidden in the background.
In this paper, we propose the unknown sniffer (UnSniffer) to find both unknown and known objects.
Firstly, the generalized object confidence (GOC) score is introduced,
which only uses \xf{known} samples for supervision and avoids improper suppression of unknowns in the background.
Significantly, such confidence score learned from \xf{known} objects can be generalized to unknown ones.
Additionally, we propose a negative energy suppression loss to further \xf{suppress} the non-object samples in the background.
Next, the best box of each unknown is hard to obtain during inference due to lacking their semantic information in training.
To solve this issue,
we introduce a graph-based determination scheme to replace hand-designed non-maximum suppression (NMS) post-processing.
Finally, we present the Unknown Object Detection Benchmark,
the first publicly benchmark that encompasses precision evaluation for \xf{unknown detection} to our knowledge.
Experiments show that our method is far better than the existing state-of-the-art methods.
Code is available at: \url{https://github.com/Went-Liang/UnSniffer}.

\end{abstract}

\section{Introduction}
\label{sec:intro}

Detecting objects with a limited number of classes in the closed-world setting \cite{ren2015faster,redmon2016you, liu2016ssd, he2017mask, lin2017feature, lin2017focal, redmon2017yolo9000, cai2018cascade, carion2020end, zhu2020deformable} has been the \xff{norm for} years.
Recently,
the popularity of autonomous driving \cite{pinggera2016lost, ramos2017detecting, dosovitskiy2017carla, 8126154, 9210191,8794279, janai2020computer, YuZhou-IJCV2016-SFVT, NIPS2012_3e313b9b, ZheLiu-AAAI2019-TANet, 2014ONLINE} has \xff{raised the bar} for object detection.
\xf{That is, the detector should \xff{detect} both known and unknown objects.
`\textbf{Known Objects}' are those that belong to pre-defined categories,
\xff{while `\textbf{Unknown Objects}'} are those that the detector has never seen during training.
Detecting unknown objects \xff{is crucial in coping} with more challenging environments,
such as autonomous driving scenes with potential hazards.}

\begin{figure}
\centering\includegraphics[width=0.97\linewidth]{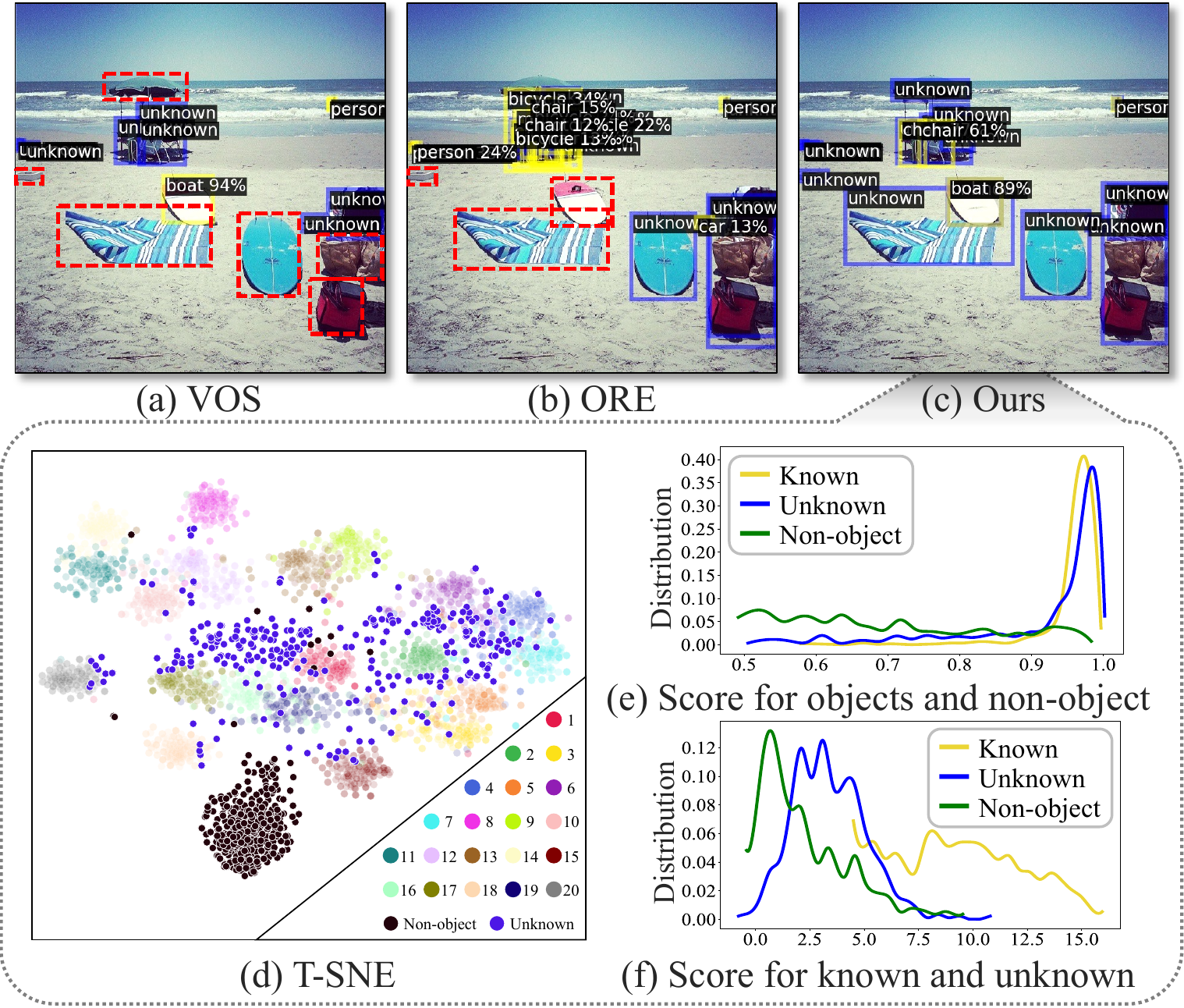}
\caption{
(a)-(c) the predicted unknown (\textcolor{blue}{blue}), known (\textcolor{yellow2}{yellow}), and missed (\textcolor{red}{red}) objects of VOS \cite{vos}, ORE \cite{owod}, and our model.
(d) t-SNE visualization of various classes' hidden vectors.
(e) score for objects and non-object (generalized object confidence).
(f) score for unknown, known, and non-object (negative energy).
}
\vspace{-10pt}
\label{fig:first_page}
\end{figure}

Since unknown objects \xf{do not have labels} in the training set,
how to learn knowledge that can be generalized to unknown classes from finite pre-defined categories is the key issue in detecting unknown objects.
In recent years,
a series of groundbreaking works have been impressive on open-set detection (OSD) \cite{vos,gal2016dropout, miller2018dropout, du2022unknown} and open-world object detection (OWOD) \cite{owod, OWDETR, yang2021objects, wu2022uc}.
Several OSD \xf{methods} have \xf{used} uncertainty measures to distinguish unknown objects from known ones.
However, they primarily focus on improving the \xf{discriminatory} power of uncertainty and tend to suppress non-objects along with many potential unknowns in the training phase.
As a result, these methods \xf{miss many} unknown objects.
Fig. \ref{fig:first_page} (a) shows that VOS \cite{vos} misses many unknown objects, such as bags, stalls and surfboards.
Furthermore,
OWOD requires generating high-quality boxes for both known and unknown objects.
ORE \cite{owod} and OW-DETR \cite{OWDETR} collect the pseudo-unknown samples by an auto-labelling step for supervision
and perform knowledge transfer from the known to the unknown by contrastive learning or foreground objectness.
\xf{But} 
the pseudo-unknown samples are unrepresentative of the unknown objects, thus limiting the model's ability to describe unknowns.
Fig. \ref{fig:first_page} (b) shows that ORE \cite{owod} mis-detects many unknown objects,
even though some are apparent.

In philosophy,
there is a concept called `\emph{Analogy}' \cite{Ribeiro2014},
which describes unfamiliar things with familiar ones.
We argue that \textit{despite being ever-changing in appearance,
the unknown objects are often visually similar to the objects \xf{of} pre-defined classes},
as observed in Fig. \ref{fig:first_page} (d).
The t-SNE visualization shows that the unknown objects tend to be \xf{among} several pre-defined classes,
while the non-objects are far away from them.
\xf{This inspires} us to express a unified concept of `object' by the proposed generalized object confidence (GOC) score learned from the known objects only.
To this end,
we first discard the background bounding boxes and only collect the object-intersected boxes for training to prevent potential unknown objects from being \xf{classified} as backgrounds.
Then, a combined loss function is designed to enforce the detector to assign relatively higher scores to boxes tightly enclosing objects.
\xf{Unlike} `objectness',
non-object boxes are not used as the negative samples for supervision.
Fig. \ref{fig:first_page} (e) shows that the GOC score distinctly separates \xf{non-objects} and \xf{`objects'}.
In addition,
we design a negative energy suppression loss on top of VOS's energy calculation \cite{vos} to further widen the gap between the non-object and the `object'.
Fig. \ref{fig:first_page} (f) shows three distinct peaks for the \xf{knowns, unknowns and non-objects.}
Next,
due to the absence of the unknown's semantic information in training,
the detector hardly determines the best bounding box by a constant threshold when the number of objects cannot be \xf{predicted} ahead of time.
In our model,
the best box determination is modelled as a graph partitioning problem,
which adaptively clusters high-score proposals into several groups and selects one from each group as the best box.

As far as we know,
the existing methods are evaluated on the COCO \cite{lin2014microsoft} and Pascal VOC benchmarks \cite{voc} that do not thoroughly label unknown objects.
Therefore, the \xf{accuracy} of unknown object detection cannot be evaluated.
Motivated by this practical need,
we propose the Unknown Object Detection Benchmark (UOD-Benchmark),
which takes the VOC's training set as the training data and contains two test sets.
(1) COCO-OOD containing objects with the unknown class only;
(2) COCO-Mix with both unknown and known objects.
They are collected from the original COCO dataset \cite{lin2014microsoft} and annotated \xf{according to} the COCO's instance labeling standard.
In addition,
the Pascal VOC testing set is employed for evaluating known object detection.

Our key contributions \xf{can be} summarized as follows:

\begin{itemize}
\item To better separate non-object and `object',
we propose the GOC score learned from known objects to express unknown objects and design the negative energy suppression to further limit non-object.

\item The graph-based box determination is designed to adaptively select the best bounding box for each object \xf{during} inference for higher unknown detection precision.

\item We propose the UOD-Benchmark containing annotation of both known and unknown objects,
enabling us to evaluate the precision of unknown detection.
We comprehensively evaluate our method on this benchmark which facilitates future use of unknown detection in real-world settings.
\end{itemize}

\section{Related Work}
\label{sec:relatedwork}

\noindent
\textbf{Open Set Classification and Detection} aim to deal with unknown samples encountered in classification or detection tasks.
Many uncertainties measuring the feature difference between unknown and known objects have been proposed,
such as OpenMax \cite{bendale2016towards}, MSP \cite{hendrycks2016baseline}, ODIN \cite{liang2017enhancing}, Mahalanobis distance \cite{denouden2018improving} and Energy \cite{liu2020energy}.
For detection, some works \cite{gal2016dropout, miller2018dropout, miller2019evaluating} used Monte Carlo dropout to generate uncertainty scores.
David \emph{et al}. \cite{hall2020probabilistic} proposed probabilistic detection quality to measure spatial and semantic uncertainty. Du \emph{et al}. \cite{vos, du2022unknown} synthesized virtual outliers to shape the decision boundary of networks and used energy as an uncertainty measure.
However,
to ensure the accuracy of detecting known objects,
they suppress both unknowns and non-objects in training,
leading to a low recall of unknowns.
In contrast,
our method aims to detect all unknown objects.

\begin{figure*}
\centering
\includegraphics[width=1\linewidth]{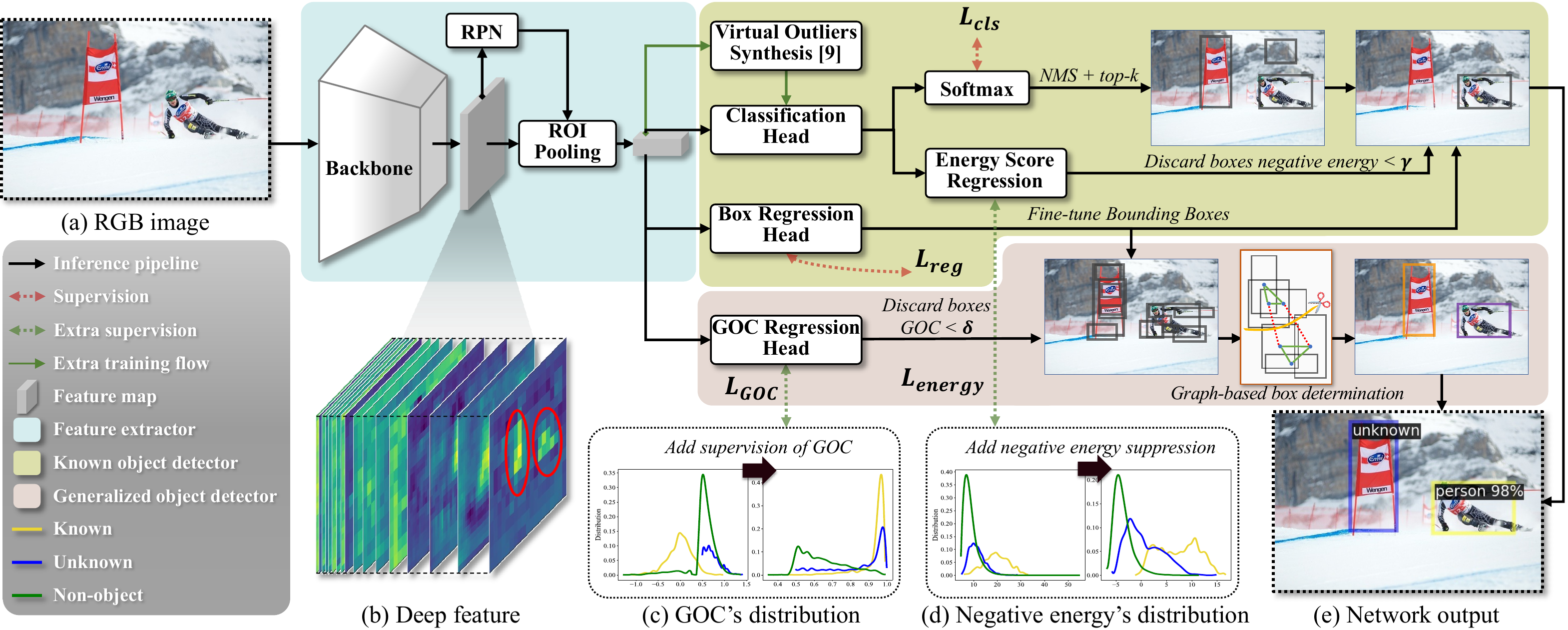}
\vspace{-16pt}
\caption{
\textbf{The framework of \net}
contains a feature extractor, a known object detector and a generalized object detector.
(a) is the input RGB image.
(b) visualizes several channels of deep features encoding the known and unknown objects at the same time,
and the red circles mark the position of the objects.
(c) shows the GOC score's distribution before and after training the GOC.
(d) shows the negative energy's distribution before and after using negative energy suppression.
(e) is the result.
}
\vspace{-10pt}
\label{fig:pipeline}
\end{figure*}

\noindent
\textbf{Open-world object detection (OWOD)} is proposed by ORE \cite{owod}.
It detects both known and unknown objects by training pseudo-labeled unknown objects and incrementally learns updated annotations of new classes.
OW-DETR \cite{OWDETR} improves performance with multi-scale self-attention and deformable receptive fields. Yang \emph{et al}. \cite{yang2021objects} introduced semantic topology to ensure that the feature representations are discriminative and consistent.
UC-OWOD \cite{wu2022uc} also classifies unknown objects to achieve better results than ORE on measures about unknown classes.
Zhao et al. \cite{rowod} correct the auto-labeled proposals by Selective Search and calibrate the over-confident activation boundary by a class-specific expelling function.
However, the auto-labeling step generates many pseudo-unknown samples that are unrepresentative of the unknowns in fact,
limiting their ability to transfer knowledge from the known to the unknown.
Thus, during inference, many non-objects are mis-detected as unknown,
leading to the low precision of unknown.
This paper aims to reduce the false positives by the proposed GOC score and a graph-based box determination scheme.

\section{Problem Formulation}

Referring to \cite{vos},
the problem of unknown detection in the setting of object detection is formulated as follows.
We have a known class set $\mathcal{K}=\{1,2,...,C\}$ and an unknown class $C+1$.
The $N$ input RGB images are denoted as $\{\mathbf{I}_1,...,\mathbf{I}_N\}$,
with corresponding labels $\{\mathbf{Y}_1,...,\mathbf{Y}_N\}$.
Each $\mathbf{Y}_i=\{\mathbf{y}_1,...,\mathbf{y}_K\}$ contains a set of object instances with $\mathbf{y}_k=[l_k,x_k,y_k,w_k,h_k]$,
where $l_k$ is the class label for a bounding box represented by $x_k,y_k,w_k,h_k$.
If $\mathbf{y}_k$ encloses a known object, $l_k\in\mathcal{K}$, otherwise $l_k = C+1$.

The model is trained on the data containing known-class objects only $\{(\mathbf{I}_n,\mathbf{Y}_n) | l_k\in\mathcal{K},\mathbf{y}_k\in\mathbf{Y}_{n}\}_{n=1}^{N^{\mathbf{train}}}$,
but tested on the data including unknown objects $\{(\mathbf{I}_n,\mathbf{Y}_n) | l_k\in\mathcal{K}\cup\{C+1\},\mathbf{y}_k\in\mathbf{Y}_{n}\}_{n=1}^{N^\mathbf{test}}$,
where $N^\mathbf{train}$ is the image number of the training set,
$N^\mathbf{test}$ for that of the test set,
and $N = N^\mathbf{test}+N^\mathbf{train}$.

\section{Method}
\label{sec:method}

We propose the unknown sniffer (UnSniffer) to find both the known and unknown objects.
The pipeline is shown in Fig. \ref{fig:pipeline}.
The RGB image $\mathbf{I}_n$
is fed into a feature extractor \cite{ren2015faster} that captures numerous object proposals $\{b_i|i\in[1,M]\}$ and their feature vectors $\{f_i|f_i\in\mathbb{R}^{1024},i\in[1,M]\}$.
Taking the feature $f_i$ as input,
we use two detectors for known and unknown objects. 

Firstly,
the generalized object detector learns the proposed generalized object confidence (GOC) score to determine whether proposal $b_i$ contains an object (See Sec. \ref{sec:MyScore}).
Then, the graph-based box determination scheme is used to cluster the high-score proposals into several groups (See Sec. \ref{sec:Ncut}).
We select the one with the highest GOC score in each group \xf{as} a set of unknown predictions.

The second one, i.e., a known object detector,
computes the class-specific probabilities
and the negative energy score \cite{vos} for $b_i$.
In addition to the classification head and box regression head commonly used in two-stage object detectors \cite{ren2015faster,he2017mask, lin2017feature,cai2018cascade},
we employ the virtual outliers synthesis \cite{vos} to learn energy scores
and remove the low-negative-energy proposals \xf{during} inference.
Unlike \cite{vos},
we employ a negative energy suppression loss to enforce the negative energy scores of non-object boxes less than zero (See Sec. \ref{sec:Energy_Suppression}).
It \xf{lowers} the feature response inside non-object boxes and boosts the discriminative power of both detectors.

Finally,
the first detector outputs the bounding box predictions of unknown class,
and the second detector gives that of known class.
We directly concatenate the two results and remove the unknown-class predictions whose IoU with any known-class prediction exceeds a constant threshold $\beta$.
Fig. \ref{fig:pipeline} (e) shows the merged result of image $\mathbf{I}_n$.

Note that the UnSniffer has two training stages,
which \xf{are} consistent with VOS \cite{vos}.
In the first stage,
we employ the training process of Faster-RCNN \cite{ren2015faster}
(the red dot arrows in Fig. \ref{fig:pipeline}),
where $L_{cls}$ and $L_{reg}$ are the losses for classification and bounding box regression, respectively.
And the second stage additionally employs the losses proposed in this paper (the green dot arrows in Fig. \ref{fig:pipeline}).

\subsection{\MyScore}
\label{sec:MyScore}

\noindent\textbf{Uncertainty \xf{Scores}} \xf{are} usually modeled as either the maximum known-class conﬁdence \xf{scores} \cite{hendrycks2016baseline,liang2017enhancing} or the entropy of the classification results \cite{liu2020energy, vos, du2022unknown}.
It can be used to distinguish unknowns from known objects according to the high uncertainty \xf{scores} of the unknown objects.
However,
the uncertainty's training phase suppresses both unknowns and non-objects,
causing the inadequate detection of unknowns.

\noindent\textbf{Objectness \xf{Scores}} \xf{are} usually used to judge whether a bounding box containing an object \cite{OLP,edgeboxes,ren2015faster},
which naturally meets the requirement of unknown object detection,
such as the foreground objectness learning in OW-DETR \cite{OWDETR} implemented by a binary classification.
However,
learning-based object proposal methods cannot avoid \xf{a} misuse of unknown samples as negative samples,
leading to low discriminative power between non-objects and unknowns.

\noindent\textbf{Generalized Object Confidence Score and Losses.}
We propose the generalized object confidence (GOC) score.
\xf{It can be used to judge} whether a proposal contains an object (including unknown and known classes),
while this capability stems from the fact that many unknowns are actually encoded by the pre-trained backbone,
as the `flag' shown in Fig. \ref{fig:pipeline}(b).

Different from uncertainty and objectness,
the GOC score is trained using only known objects and can be generalized to unknown objects.
Specifically,
the GOC regression head that is composed of a linear transformation, denoted as $\Phi$, is used to compute the GOC score $\Phi(f_i)$ for a given proposal's feature $f_i$.
In the training phase,
given an image-label pair $(\mathbf{I}_n,\mathbf{Y}_n)$,
the region proposal network is firstly used to extract numerous proposals $B_n=\{b_i|i\in[1,M]\}$ from image $\mathbf{I}_n$.
And we define the intersection over the predicted bounding box ($IoP$) and \xf{the} intersection over the correct bounding box ($IoC$) for collecting training samples from $B_n$ as follows:
\vspace{-6pt}
\begin{equation}
IoP(b_i, \mathbf{y}_k)=\frac{|b_i \cap \mathbf{y}_k|}{|b_i|}, \quad IoC(b_i, \mathbf{y}_k)=\frac{|b_i \cap \mathbf{y}_k|}{|\mathbf{y}_k|}
\label{eq:IoPIoC}
\end{equation}
where $\mathbf{y}_k$ is the $k$-th instance's bounding box in $\mathbf{Y}_n\!\!\!=\!\!\!\{\mathbf{y}_1,...,\mathbf{y}_K\}$.
Subsequently, for each proposal $b_i$ in $B_n$,
we find the object instance \xf{that has} the maximum IoU with $b_i$.
And the proposals enclosing the same object are assigned to the same group,
obtaining $K$ groups of proposals: $B^1_n,B^2_n,..,B^K_n$.
Then,
we divide the proposals of $B^k_n$ into \textbf{c}omplete-object, \textbf{p}artial-object, \textbf{o}versized, and \textbf{n}on-object according to IoU, IoP, and IoC,
as shown in Fig. \ref{fig:box}:
\begin{equation}
\centering
\begin{split}
\!\!\!\mathbf{B}^{k,\mathbf{c}}_{n} \!&=\!\! \{b_i\!\in\! B^k_n | IoU(b_i, \mathbf{y}_k) \geq e_2\}\! \\
\!\!\!\mathbf{B}^{k,\mathbf{p}}_{n} \!&=\!\! \{b_i\!\in\! B^k_n | e_1 \!\!\leq\!\! IoU(b_i, \mathbf{y}_k)\!\!<\!\!e_2, IoP(b_i, \mathbf{y}_k) {\geq} {\rho}\}\! \\
\!\!\!\mathbf{B}^{k,\mathbf{o}}_{n} \!&=\!\! \{b_i\!\in\! B^k_n | e_1 \!\!\leq\!\! IoU(b_i, \mathbf{y}_k)\!\!<\!\!e_2, IoC(b_i, \mathbf{y}_k) {\geq} {\rho}\}\! \\
\!\!\!\mathbf{B}^{k,\mathbf{n}}_{n} \!&=\!\! \{b_i\!\in\! B^k_n | IoU(b_i, \mathbf{y}_k) < e_1\}\! \\
\end{split}
\label{eq:S_plsn}
\end{equation}
where $e_1, e_2, \rho$ are the constant thresholds,
as shown in Fig. \ref{fig:box}.
In order to prevent the potential \ood from being \xf{treated} as background during training,
we only use the first three groups in Eq. \ref{eq:S_plsn} to train the module $\Phi$ \xf{with} three losses.
In the first loss,
the GOC scores of complete-object bounding boxes are pushed towards one:

\begin{figure}
\centering
\includegraphics[width=1\linewidth]{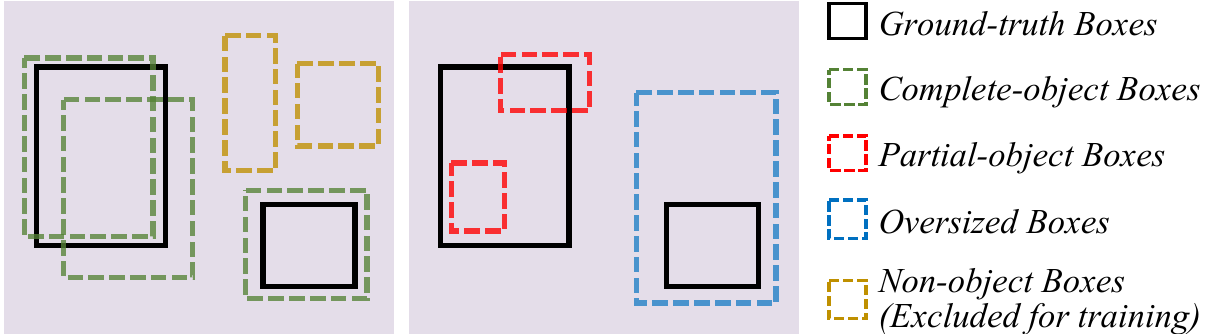}
\caption{The sample definition in GOC supervision.}
\vspace{-10pt}
\label{fig:box}
\end{figure}
\begin{equation}
\vspace{-6pt}
 L_{pos}\!\!=\!\! \frac{1}{K}\!\sum_{k\in[1,K]}\! \frac{1}{|B^{k,\mathbf{c}}_{n}|}\sum_{b_i\in \mathbf{B}^{k,\mathbf{c}}_{n}} \big(\Phi(f_i)- 1\big)^2
\label{eq:posloss}
\end{equation}
Then, due to the lack of clear criteria measuring GOC scores of partial-object or oversized boxes,
we suppress their GOC scores to below a constant $\delta$:
\begin{figure*}
\centering
\includegraphics[width=0.95\linewidth]{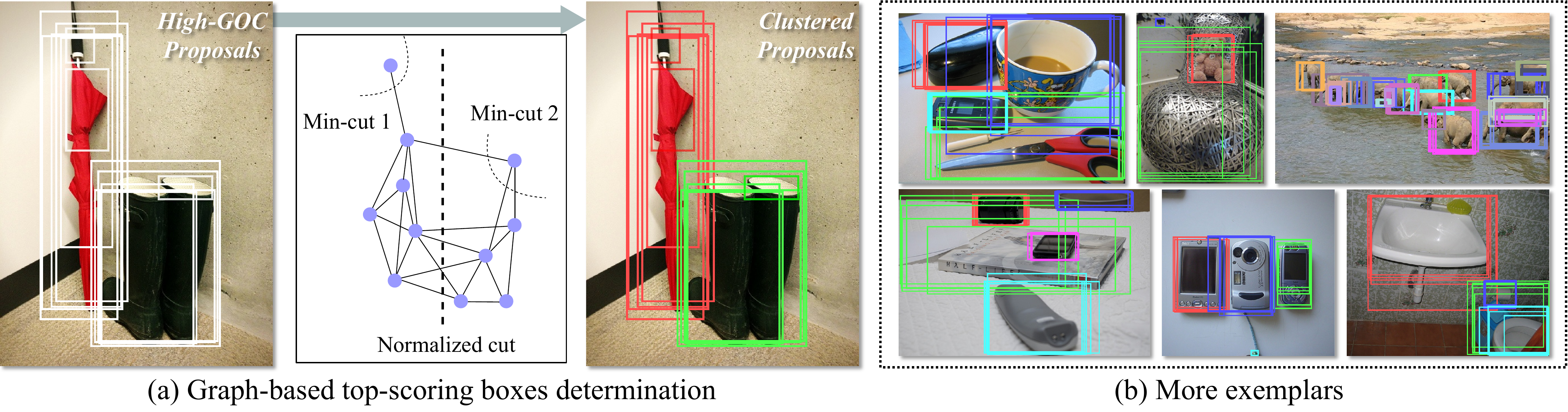}
\vspace{-5pt}
\caption{The illustration and more examples for the graph-based top-scoring box determination.
The white rectangles denote the proposals with top GOC scores in the image.
For other rectangles,
each group of proposals is represented by the same color.}
\vspace{-10pt}
\label{fig:ncut}
\end{figure*}

{\setlength\abovedisplayskip{0.18cm}
\setlength\belowdisplayskip{0.18cm}
\begin{equation}
\!\!L_{neg}\! =\!\! \frac{1}{K}\!\!\sum_{k\in[1,K]}\!\frac{1}{|B^{k,\mathbf{po}}_{n}|}\!\sum_{b_i\in {\mathbf{B}^{k,\mathbf{po}}_{n}}} \!\max\!\big(0, \Phi(f_i) \!-\! \delta\big)
\label{eq:negloss}
\end{equation}}where $B^{k,\mathbf{po}}_{n}=B^{k,\mathbf{p}}_{n}\cup B^{k,\mathbf{o}}_{n}$.
Next,
we improve the model's ability to capture a box enclosing an object more entirely by a contrastive loss,
which compares two boxes in $\mathbf{B}^{k,\mathbf{c}}_{n}$:
{\setlength\abovedisplayskip{0.18cm}
\setlength\belowdisplayskip{0.18cm}
\begin{equation}
\begin{split}
\!\!\!\!\!L_{con}\!\!&=\!\!\frac{1}{K}\!\!\sum_{\!k\in[1,K]}\! \!\Bigg\lfloor\!\frac{2}{|\mathbf{B}^{k,\mathbf{c}}_{n}|}\!\Bigg\rfloor\!\!\!\!\!\!\!\!\sum_{\quad b_i\!,b_j\in\mathbf{B}^{k,\mathbf{c}}_{n}} \!\!\!\!\!\!\!\!\!\!\max\!\big(0,\!\frac{\Phi(\!f_i\!)\!-\!\Phi(\!f_j\!)}{\alpha}\!+\!\zeta\big)\\
\end{split}
\label{eq:contrastive}
\end{equation}}where $\alpha = 1$ when $IoU(b_j, \mathbf{y}_k) > IoU(b_i, \mathbf{y}_k)$,
otherwise $\alpha = -1$.
$\zeta$ is a tiny constant that is set to 0.01,
and $i\ne j$.
Finally, the total GOC loss is formulated as:
{\setlength\abovedisplayskip{0.18cm}
\setlength\belowdisplayskip{0.18cm}
\begin{equation}
L_{GOC} = L_{neg} + L_{pos} + L_{con}
\label{eq:CIloss}
\end{equation}}

Since our training process does not utilize the sample of the background area,
the GOC scores of non-object bounding boxes would not be affected greatly.
On the contrary, the GOC scores of both unknown and known objects are pushed to a high score.
As shown in Fig. \ref{fig:pipeline} (c),
when the GOC is not supervised,
unknown boxes have the same output as non-object boxes,
but it changes dramatically when the GOC regression head is supervised by $L_{GOC}$.

\subsection{Graph-based Top-scoring Box Determination}
\label{sec:Ncut}
By using the GOC score for ranking proposals during inference,
we obtain the proposals where the objects are most likely to be.
However,
the traditional post-processing mechanism, i.e. using NMS and outputting top-$k$ highest results, is inappropriate to determine the unknown prediction,
as the number of objects cannot be prophesied at ahead of time.

To address this issue,
we perform the top-scoring box determination as a graph partitioning problem,
which adaptively finds the best bounding box for each object,
as shown in Fig. \ref{fig:ncut}(a).
Specifically,
during the inference,
given a set of proposals $\{b_i|i\in[1,M]\}$ and their GOC scores $\{\Phi(f_i)|i\in[1,M]\}$ for image $\mathbf{I}_n$,
we construct a weighted undirected graph $G=(V,E)$.
Each node in set $V$ represents an object proposal $b_i$,
and each edge in set $E$ is formed by the IoU between both ends of this edge, i.e. $IoU(b_i,b_j)$.
As shown in Fig. \ref{fig:ncut}(a),
considering that some of the proposals may only cover part of the objects,
we employ the recursive two-way normalized cut algorithm \cite{ncut} to decompose the entire graph $G$ into several sub-graphs iteratively,
which is terminated until the NCut value \cite{ncut} of a sub-graph is lower than a threshold $\varepsilon$,
where $\varepsilon$ is determined by a threshold selection method in Sec. \ref{sec:imp}.
Finally, the top-$1$ GOC proposal of each sub-graph is taken as the prediction.

It can be seen from Fig. \ref{fig:ncut}(b),
even if only IoU is used as the measurement of edge in graph $G$,
our model still performs considerable proposal clustering and avoids outlier proposals as an independent group.
Especially in the upper right prediction of Fig. \ref{fig:ncut}(b),
almost every elephant gets an independent group of proposals.

\subsection{Negative Energy Suppression for Non-object}
\label{sec:Energy_Suppression}

Referring to VOS \cite{vos},
the energy score is employed to distinguish unknown objects from known ones,
as shown in Fig. \ref{fig:pipeline}.
For proposal $b_i$,
the energy score is formulated as the negative of the weighted sum of this proposal's logit output in exponential space:
\begin{equation}
E(b_i)=-\log \sum\nolimits_{c\in[1,C]} \mathbf{w}_{c} \cdot \exp ^{\mathbf{f}_{c}}
\label{eq:energy}
\vspace{-5pt}
\end{equation}
where $\mathbf{f}_{c}$ is the logit output for class $c$ in the classification head,
$C$ is the number of the know classes,
and $\mathbf{w}_{c}$ is the learnable parameter for alleviating the class imbalance.
The proposals with higher negative energy scores are treated as known predictions,
\xf{whereas} others are unknown predictions.
However, due to insufficient training,
some non-object proposals gain such high negative energy scores that they are indistinguishable from objects, as shown in the left plot of Fig. \ref{fig:pipeline}(d).
To address this issue,
we propose negative energy suppression to further reduce the negative energy scores of non-object proposals.
Specifically,
we observe that most non-object boxes have lower negative energy scores than \xf{those} of the known and unknown classes,
which motivates us to design a suppression loss to constraint $T$ proposals with the lowest negative energy scores:
\vspace{-5pt}
\begin{equation}
L_{suppression} = \frac{1}{T} \sum\nolimits_{i\in[1,T]} \max(0, -E(b_i))
\label{eq:loss_suppression}
\vspace{-5pt}
\end{equation}

The overall energy loss consists of our proposed $L_{suppression}$ and $L_{uncertainty}$ defined by VOS \cite{vos}:
\vspace{-5pt}
\begin{equation}
L_{energy} = L_{suppression} + L_{uncertainty}
\label{eq:loss_energy}
\vspace{-5pt}
\end{equation}

As shown in the right plot of Fig. \ref{fig:pipeline}(d),
after training with loss $L_{energy}$,
the negative energy distribution of the non-object is significantly different from that of the unknown,
indicating that the non-objects are indeed suppressed.
In addition,
by using Eq. \ref{eq:loss_suppression},
the feature responses of non-object bounding boxes are reduced simultaneously,
which further widens the GOC difference between non-object and object.
It can be proved by the fact that the high GOC scores of unknowns in Fig. \ref{fig:first_page}(e) (with $L_{suppression}$) are more than that in the right plot of Fig. \ref{fig:pipeline}(c) (without $L_{suppression}$).
In addition,
the ablation studies of Sec. \ref{sec:Analysis} demonstrate that this approach improves the detection precision of the model.

\section{Unknown Object Detection Benchmark}
\label{sec:experimental}

\subsection{Datasets}
\begin{figure}[t!]
\includegraphics[width=1\linewidth]{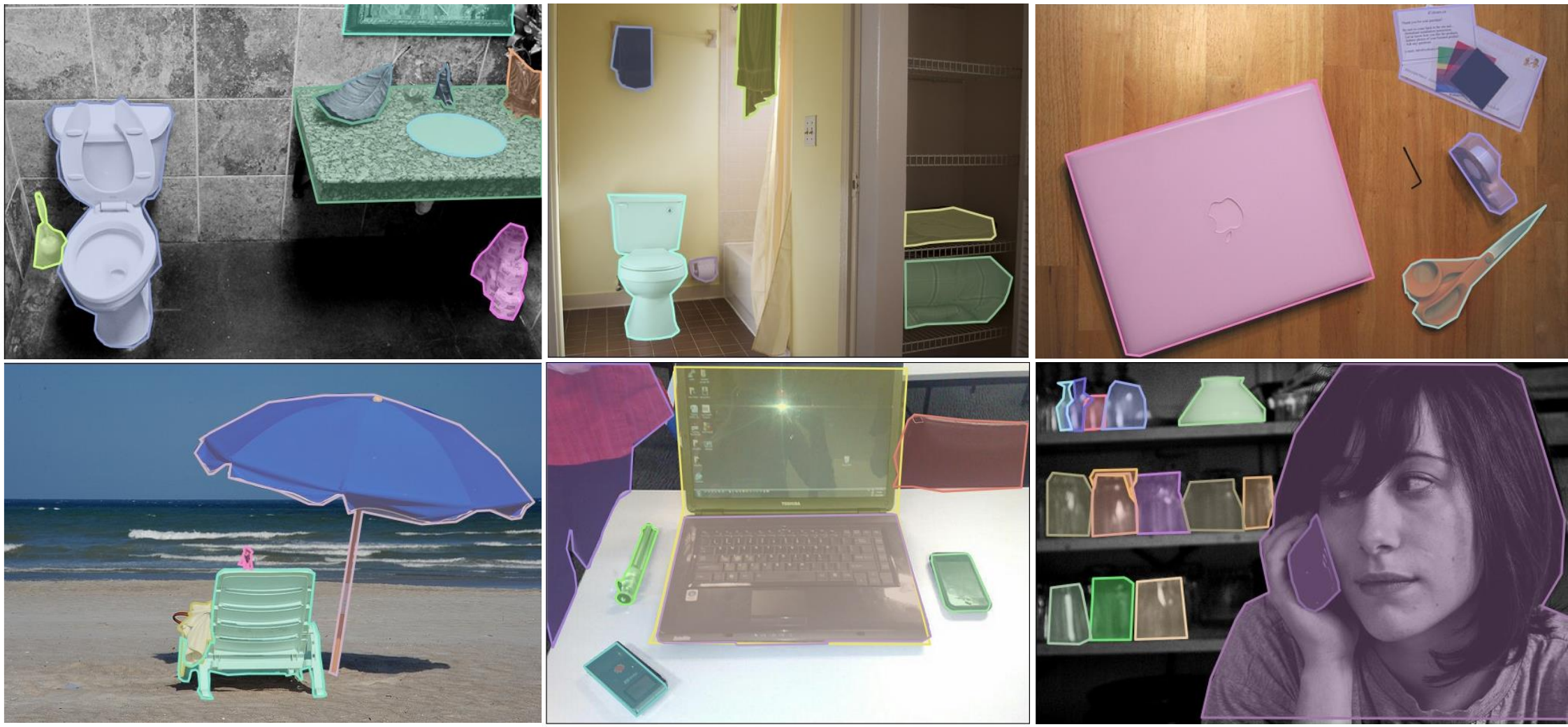}
\caption{Annotated samples in COCO-OOD and COCO-Mix.}
\vspace{-8pt}
\label{fig:dataset}
\end{figure}

In the proposed UOD-Benchmark,
we refer to \cite{vos,owod} 
\xf{and use} the Pascal VOC dataset \cite{voc} as the training data that contains annotations of 20 object categories.
For testing,
since the MS-COCO dataset \cite{lin2014microsoft} extends the PASCAL VOC categories to 80 object categories,
we naturally employ MS-COCO \xf{to evaluate} unknown objects.
However, MS-COCO does not thoroughly label potential unknown objects in images.
To address this issue,
we propose two datasets, i.e., COCO-OOD and COCO-Mixed, which fully \xf{label} the unknown objects.
Firstly,
according to the definition of objects in COCO \cite{lin2014microsoft},
i.e. ``objects are individual instances that can be easily labelled (person, chair, car)'',
we hand-pick more than a thousand images that have no area confused with this definition.
Secondly,
several master students are asked to mark the object regions they got at first glance by drawing polygons,
referring to the object definition above.
As shown in Fig. \ref{fig:dataset},
we label almost every object in the selected images with fine-grained annotation.
Finally, we have two datasets both for testing as follows:

\noindent\textbf{COCO-OOD} dataset \xf{contains only} unknown categories,
consisting of 504 images with ﬁne-grained annotations of 1655 unknown objects.
All annotations consist of original annotations in COCO and the augmented annotations on the basis of the COCO definition.


\noindent\textbf{COCO-Mixed} dataset includes 897 images with annotations of both known and unknown categories. 
It contains 2533 unknown objects and 2658 known objects, with original COCO annotations used as labels for known objects. 
Unambiguous unlabeled objects are also annotated. 
The dataset is more challenging to evaluate due to the images containing more object instances with complex categories and concentrated locations.

\begin{table}[t!]
\centering
\begin{tabular}{c|c|c c}
\toprule
Datasets               & Images & Known        & Unknown\\ \midrule
VOC-Pretest            & 200    & 5.09         & 0      \\
VOC-Test               & 4952   & 3.02         & 0      \\
COCO-OOD$^\clubsuit$   & 504    & 0            & 3.28   \\ 
COCO-Mixed$^\clubsuit$ & 897    & 2.96         & 2.82   \\\bottomrule
\end{tabular}
\caption{\textbf{The Statistics of datasets} \xf{that include} the number of images, the average number of known and unknown instances per image.
$\clubsuit$ denotes the augmented datasets.}
\vspace{-7pt}
\label{tab:datasets}
\end{table}

In addition, we employ the test set of the Pascal VOC dataset to evaluate the accuracy of known object detection.
The statistics of these test datasets are placed in Table \ref{tab:datasets}. Fig. \ref{fig:dataset} shows some fully annotated images in COCO-OOD and COCO-Mixed.
Note that the VOC-Pretest is used to set a threshold,
mentioned in Sec .\ref{sec:imp}.

\subsection{Evaluation Metrics}

To evaluate the performance of known object detection,
we employ a prevalent metric,
i.e., mean Average Precision (mAP) \cite{vos}.
As for the unknown detection performance,
assuming that $TP_u$ denotes the true positive proposals of unknown classes,
$FN_u$ for false negative proposals,
and $FP_u$ for false positive proposals,
five metrics are employed:
\begin{itemize}
    \item The Unknown Average Precision (U-AP) is used with reference to the conventional object detection \cite{voc}.
    \item The Recall Rate (U-REC) and Precision Rate of Unknown (U-PRE) are defined \xf{as follows} $U\mbox{-}REC = \frac{TP_u}{TP_u+FN_u}, U\mbox{-}PRE = \frac{TP_u}{TP_u+FP_u}$.
    \item For a comprehensive comparison,
    we report the Unknown F1-Score defined as the harmonic mean of U-PRE and U-REC: $U\mbox{-}F1 = \frac{2 \times U\mbox{-}PRE \times U\mbox{-}REC}{U\mbox{-}PRE + U\mbox{-}REC}$.
    \item The Absolute Open-Set Error (A-OSE) \cite{miller2018dropout} is also employed to report the number count of unknown objects that are wrongly classiﬁed as any known classes.
    \item Wilderness Impact (WI) \cite{dhamija2020overlooked} are defined as $WI = \frac{Precision\ in\ closed{-}set}{Precision\ in\ open{-}set} - 1$ to characterize the case that unknown objects are confused with the known.
\end{itemize}

\xf{Note} that we measure mAP over different IoU thresholds from 0.5 to 0.95.
Other metrics, such as U-REC, U-PRE, etc., are measured at the IoU threshold of 0.5. WI is measured at a recall rate of 0.8.

\section{Experiment}
\subsection{Implementation Details}
\label{sec:imp}
We use the Detectron2 \cite{wu2019detectron2} library and employ a ResNet-50 \cite{he2016deep} backbone.
$\delta$ in Eq. \ref{eq:negloss} is empirically set to 0.5.
$\beta$ is set to 0.98.
We set the thresholds $e_1$ in Eq. \ref{eq:S_plsn} to 0.0,
and $e_2, \rho$ to 0.5 by parameter experiments.
The proposal number $T$ is set to 100 in Eq.\ref{eq:loss_suppression}.
We train the model for a total of 18,000 iterations.
The starting iteration of our second stage is 12000,
which is consistent with VOS \cite{vos}.

\begin{table*}
\centering
\begin{threeparttable}
\setlength{\tabcolsep}{1.3mm}{
\resizebox{\linewidth}{!}{
\begin{tabular}{c|l|c|cccc|ccccccc}
\toprule
\multirow{2}{1cm}{Groups}&\multirow{2}{2cm}{Methods} & VOC-Test & \multicolumn{4}{c|}{COCO-OOD} & \multicolumn{7}{c}{COCO-Mix} \\ 
\cline{3-14} 
& & mAP & U-AP & U-F1 & U-PRE & U-REC & mAP & U-AP & U-F1 & U-PRE & U-REC & AOSE & WI \\
\hline
\ding{172} &Faster-RCNN \cite{ren2015faster} & 0.483 & - & - & - & - & - & - & - & - & - & - & -\\
 \hline
\multirow{3}{1cm}{\centering\ding{173}} & MSP\cite{hendrycks2016baseline} & 0.470 & 0.213 & 0.314 & 0.279 & 0.359 & 0.364 & 0.055 & 0.169 & 0.190 & 0.153 & 588 & 0.135\\
&Mahalanobis \cite{denouden2018improving} &0.447 &0.129 &0.271  & 0.309 &0.241 & 0.351 & 0.051 & 0.149 & 0.207 & 0.116 & 604 & 0.165 \\
&Energy score \cite{liu2020energy} &\underline{0.474} &0.213 &0.308  & 0.260 &0.377 & 0.364 & 0.048 & 0.169 & 0.167 & 0.171 & 470 & 0.137\\
\hline
\multirow{2}{1cm}{\centering\ding{174}} &OW-DETR \cite{OWDETR} &0.420 &0.033  &0.056 &0.030 &0.380 &\textbf{0.414}  &0.007  &0.025 &0.014  &0.161  &569 & \textbf{0.086}\\
&ORE \cite{owod} & 0.243 & \underline{0.214} &0.255 & 0.153 & \textbf{0.782} & 0.213 & \underline{0.140} & \underline{0.175} & 0.103 & \textbf{0.592} & 485 &\underline{0.089}\\
\hline
\multirow{2}{1cm}{\centering\ding{175}} &VOS$^{1}$ \cite{vos} & \textbf{0.485} & 0.135 &0.196 & \underline{0.342} & 0.137 & \underline{0.377} & 0.040 &0.101 & \textbf{0.262} & 0.062 & 640 & 0.152\\
&VOS$^{2}$ \cite{vos} & 0.469 & 0.205 &\underline{0.317} & 0.291 & 0.348 & 0.364 & 0.051 & 0.172 & 0.184 & 0.163 & \underline{409} &0.124 \\
\hline
\ding{176}&Ours & 0.464 & \textbf{0.454} &\textbf{0.479} & \textbf{0.433} & \underline{0.535} & 0.359 & \textbf{0.150} &\textbf{0.287} & \underline{0.222} & \underline{0.409} & \textbf{398} & 0.175\\
\bottomrule
\end{tabular}}}
\vspace{-0.5em}
\caption{Comparisons with the traditional detector \ding{172} and detectors using open-set classification \ding{173},
open-world object detection \ding{174},
and open-set detection \ding{175} methods. 
VOS\tnote{1} means using the threshold in the official repository,
calculated on the BDD100K dataset \cite{yu2020bdd100k}.
VOS\tnote{2} means using the threshold computed on the COCO-OOD dataset by the official code.
Best results are in bold,
second best are underlined.}
\begin{tablenotes}
\scriptsize
\item[1] https://github.com/deeplearning-wisc/vos/issues/26
\item[2] https://github.com/deeplearning-wisc/vos/issues/13
\end{tablenotes}
\end{threeparttable}
\label{tab:all_result}
\vspace{-0.5em}
\end{table*}

\noindent\textbf{Determining inference thresholds in pretest mode.}
Both VOS \cite{vos} and OWOD \cite{owod} determine \xf{the} threshold before inference,
but they bring in unknown data in an implicit or explicit way when computing the thresholds.
Therefore, we introduce a pretest operation before inference,
which \xf{selects 200 images from} the training set that do not contain any potential unknown objects for threshold determination.
The first row of Table \ref{tab:datasets} shows the statistics of the pretest dataset.
In the pretest mode,
we firstly obtain the negative energy score of the proposals predicted from the pretest dataset,
and set the threshold $\gamma$ such that 95\% of predicted proposals have a negative energy score greater than it.
Then,
for the graph-based box determination,
we set 10 thresholds of NCut value equally spaced from 0 to 1,
and choose the threshold when the AP of known objects is the largest as $\varepsilon$.

\begin{table}
\small
\centering
\setlength{\tabcolsep}{1.2mm}{
\begin{tabular}{c|ccc|cccc}
\toprule
Row & GOC & NES & GBD &U-AP &U-F1 & U-PRE & U-REC \\
\midrule
1 & $\times$ & $\times$  & $\times$ &0.066 &0.050 &0.026 &0.808 \\
2 & $\times$ & $\times$  & \checked &0.250 &0.434 &0.395 &0.481 \\
3 & $\times$ & \checked  & $\times$ &0.442 &0.054 &0.028 &0.861 \\
4 & \checked & $\times$  & $\times$ &0.479 &0.323 &0.215 &0.646 \\
\midrule
5 & $\times$ & \checked  & \checked &0.409 &\underline{0.467}&\textbf{0.437} &0.502 \\
6 & \checked & $\times$  & \checked &\underline{0.455} &0.454 &0.399 &0.528 \\
7 & \checked & \checked  & $\times$ & \textbf{0.474} & 0.342 & 0.234 & \textbf{0.632} \\
8 & \checked & \checked  & \checked &0.454 &\textbf{0.479} &\underline{0.433}  &\underline{0.535} \\
\bottomrule
\end{tabular}
}
\vspace{-0.5em}
\caption{
\textbf{Ablation studies on COCO-OOD}.
GOC, NES and GBD refer to `generalized object confidence', `negative energy suppression' and `graph-based box determination', respectively.
If `GBD' is $\times$, we use NMS+top-$k$ with \xf{the} threshold of the known detector.}
\vspace{-1em}
\label{tab:Ablation}
\end{table}

\subsection{Results}

\noindent\textbf{Quantitative Analysis.}
In Table \ref{tab:all_result},
we show \net's result on the UOD-Benchmark,
along with the results of MSP \cite{hendrycks2016baseline}, Mahalanobis \cite{denouden2018improving}, Energy score \cite{liu2020energy}, ORE \cite{owod}, OW-DETR \cite{OWDETR} and VOS \cite{vos}.
Note that OW-DETR is based on Deformable DETR \cite{zhu2020deformable} with a stronger discriminative power,
while other methods use Faster-RCNN.
Since U-PRE or U-REC cannot independently reflect the model's performance,
we mainly employ U-AP, U-F1, AOSE and WI.
Observably,
on the COCO-OOD dataset, the U-AP of UnSniffer outperforms the 2$^{nd}$ result by more than twice, and our U-F1 is 16.2\% higher than the 2$^{nd}$ result,
at the cost of a 1.9\% drop in mAP on VOC compared to Faster-RCNN \cite{ren2015faster}.
On the COCO-Mix dataset,
the UnSniffer still holds the lead in both U-AP and U-F1,
which are 1\% and 11.2\% higher than the 2$^{nd}$ results, respectively.
Those comparisons demonstrate that UnSniffer outperforms the existing methods in unknown object detection,
which \xf{is owed to} to our GOC learning the overall confidence of objects from finite known objects.
Furthermore, UnSniffer has the smallest AOSE (398) but the largest WI (0.175),
which can be explained by the inverse relationship between WI and the count of known objects misclassified as an incorrect class.
More details are illustrated in the supplementary material.

\begin{figure}
\centering
\includegraphics[width=1\linewidth]{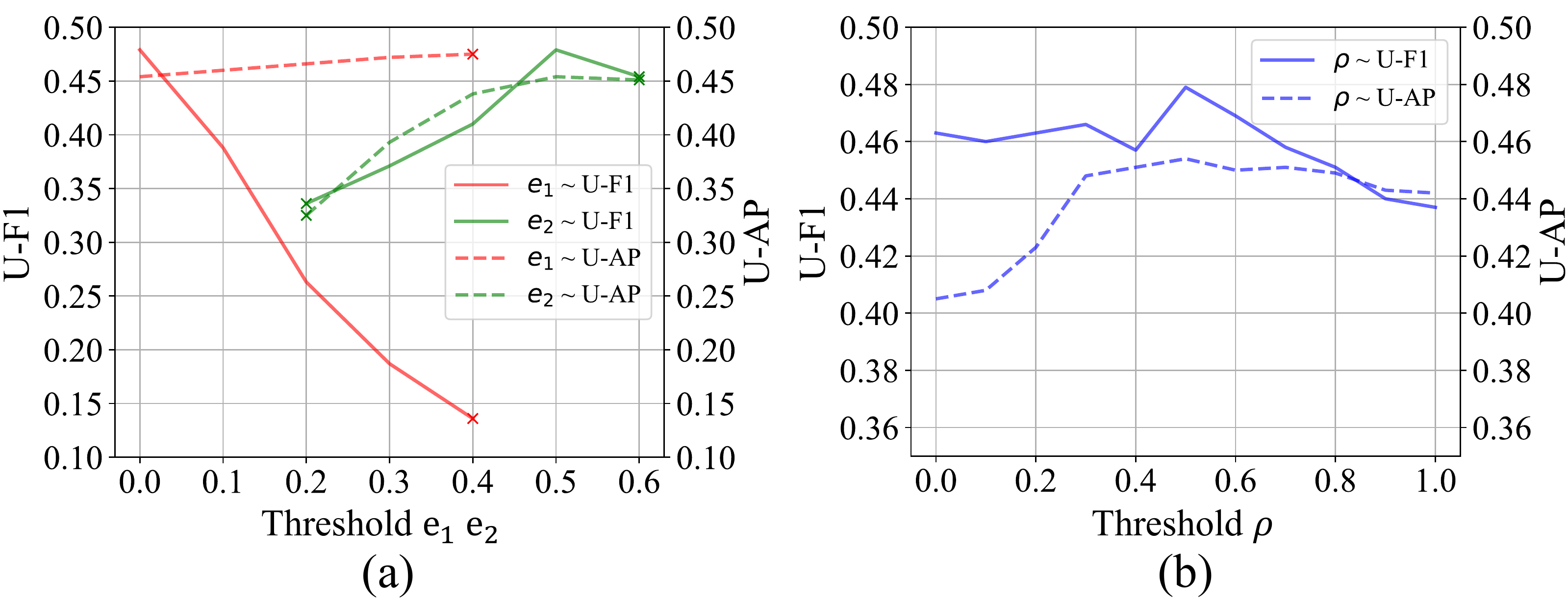}
\vspace{-2em}
\caption{
\textbf{Sensitivity analysis on (a) thresholds $e_1, e_2$, and (b) threshold $\rho$}.
$\times$ indicates the failed training outside this threshold.
}
\vspace{-0.6em}
\label{fig:samplingandstrategies}
\end{figure}

\begin{figure*}[t!]
\centering
\includegraphics[width=1.0\textwidth]{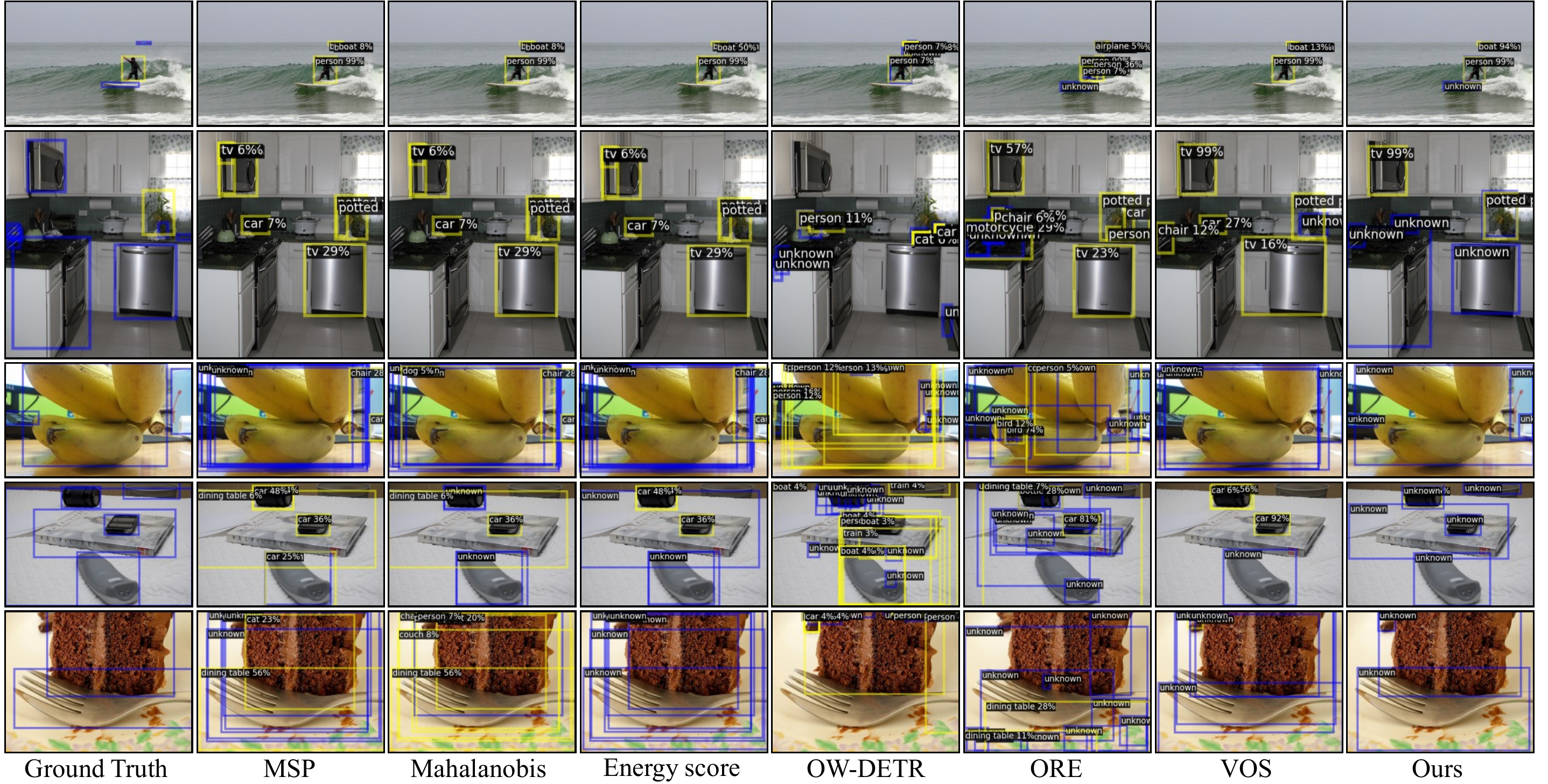}
\vspace{-1.5em}
\caption{\textbf{Example results on COCO-Mix (first two rows) and COCO-OOD datasets (last three rows)}.
1$^{st}$ column: ground truth;
2$^{nd}$-8$^{th}$ columns:
MSP \cite{hendrycks2016baseline}, Mahalanobis \cite{denouden2018improving}, Energy score \cite{liu2020energy}, OW-DETR \cite{OWDETR}, ORE \cite{owod}, VOS \cite{vos} (with threshold computed on COCO-OOD dataset), and our method.
The detections are overlaid on the known (\textcolor{yellow2}{yellow}) and unknown (\textcolor{blue}{blue}) class objects.
Since ORE and OW-DETR generate too many results,
we only draw the top-$10$ boxes for each image \xf{, and} other methods draw all predicted boxes.}
\label{fig:Qualitative}
\vspace{-1.1em}
\end{figure*}

\noindent\textbf{Qualitative Analysis.}
\label{sec:Qualitative}
Fig. \ref{fig:Qualitative} visualizes the results of different methods on example images of the COCO-Mix (first two rows) and COCO-OOD dataset (last three rows).
It can be seen that VOS \cite{vos}, MSP \cite{hendrycks2016baseline}, Mahalanobis distance \cite{denouden2018improving}, and Energy score \cite{liu2020energy} miss many objects of the unknown class,
such as the surfboards in the $1^{st}$ image,
the keyboard and water cup in the $3^{rd}$ image,
the CD case in the $4^{th}$ image.
Their failure is due to the suppression of unknown objects in training.
For the OWOD methods,
ORE and OW-DETR generate too many predicted object boxes on almost all images.
Most predictions are false positives.
In contrast,
UnSniffer does not miss any unknown objects because we give reasonable GOC scores to all unknown objects.
Moreover, using graph-based box determination,
UnSniffer reliably predicts a single bounding box for each object even if two or more objects overlap (See the 4$^{th}$ row).
More results are available in the supplementary material.

\begin{figure}
\centering
\includegraphics[width=1\linewidth]{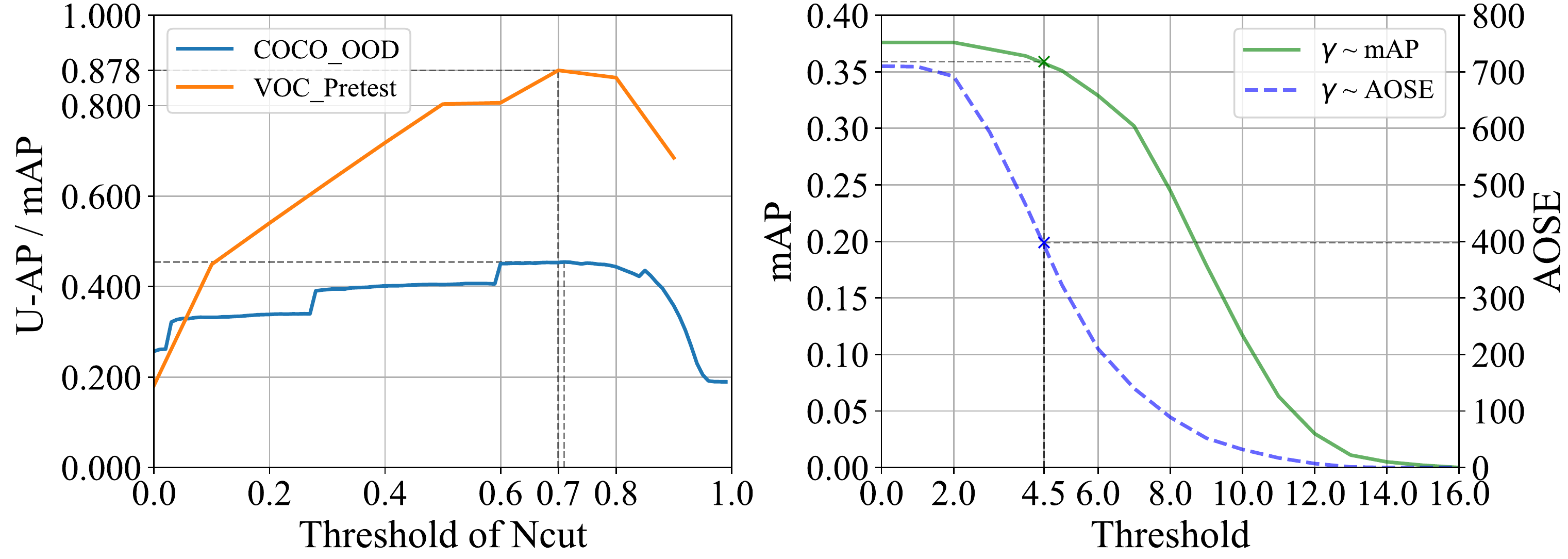}
\vspace{-2em}
\caption{
(a) Comparison between the thresholds $\varepsilon$ determined in the pretest set and the COCO-OOD dataset. 
(b) Validation of the threshold $\gamma$ computed in pretest mode on the COCO-Mix dataset.
}
\vspace{-14pt}
\label{fig:pretest}
\end{figure}

\subsection{Discussions and Analysis}
\label{sec:Analysis}
\noindent\textbf{Ablation Study.}
To investigate the contribution of each component in \net,
we design ablation experiments in Table \ref{tab:Ablation}.
Since the softmax in known object detector fails to provide reasonable confidences for unknowns,
when the 'GOC' is $\times$,
we employ the negative energy score for ranking the unknown proposals.
Comparing the 1$^{st}$ and 2$^{nd}$ rows,
it can be seen that the GBD outperforms NMS,
and GBD plays a key role in improving detection precision.
Comparing the 1$^{st}$ and 4$^{th}$ rows,
obviously,
GOC helps to increase U-AP by 41.3\%,
which shows that the GOC score well measures the probability that a proposal belongs to an object.
Using both NES and GBD (5$^{th}$ row) results in good U-PRE performance,
indicating the effectiveness of NES in reducing false-positive unknown objects.
Finally, using all modules achieves high U-PRE and U-REC simultaneously.

\noindent\textbf{Parameter sensitivity analysis of GOC sampling.}
We adjust the thresholds $e_1, e_2, \rho$ in Eq. \ref{eq:S_plsn}, respectively.
And the results on the COCO-OOD dataset are shown in Fig. \ref{fig:samplingandstrategies}.
The best result is achieved when $e_1=0$, $e_2=0.5$, and $\rho=0.5$.
Note that, when $e_1$ exceeds 0.4,
the network cannot be successfully trained.
When $e_2$ is lower than 0.2,
the training cannot converge,
which means that the contrastive loss $L_{con}$ is unsuitable for supervising too many samples.

\noindent\textbf{Effectiveness of pretest-based threshold determination.}
As shown in Fig. \ref{fig:pretest} (a),
the optimal threshold of $\varepsilon$ determined by pretest data is almost the same as that determined by traversing the COCO-OOD dataset.
Fig. \ref{fig:pretest} (b) shows the curve of mAP and AOSE when using different $\gamma$.
Note that $\gamma$ is determined to be 4.5 in the pretest data.
When $\gamma$ is equal to 4.5,
the mAP loss of known objects is small,
but the AOSE is largely reduced.
It shows that UnSniffer largely retains the ability to detect known objects,
meanwhile effectively alleviating the false detection of the unknown objects as a known class.
These comparisons prove that the pretest mode is suitable for determining the threshold of an unknown object detector.
And it only uses part of the training data without any risk of leaking test data.

\section{Conclusion}
\label{sec:conclusion}
To meet the real-world requirements for perceiving known and unknown objects,
the \net is designed.
Firstly,
we design the GOC score that reliably measures the probability of a box that contains an object.
Then, we model the top-scoring box determination as graph partitioning to obtain the best box for each object.
Thirdly,
the proposed negative energy suppression further limits the non-object boxes.
Finally, we introduce the UOD-Benchmark to more comprehensively evaluate the real-world usability of the model.
We hope our work inspires future research on unknown object detection in real-world settings.

{\small
\bibliographystyle{ieee_fullname}
\bibliography{UnSniffer_CameraReady_v2}

\begin{thebibliography}{10}\itemsep=-1pt

\bibitem{bendale2016towards}
Abhijit Bendale and Terrance~E Boult.
\newblock Towards open set deep networks.
\newblock In {\em IEEE Conference on Computer Vision and Pattern Recognition
  (CVPR)}, 2016.

\bibitem{cai2018cascade}
Zhaowei Cai and Nuno Vasconcelos.
\newblock Cascade r-cnn: Delving into high quality object detection.
\newblock In {\em IEEE Conference on Computer Vision and Pattern Recognition
  (CVPR)}, 2018.

\bibitem{carion2020end}
Nicolas Carion, Francisco Massa, Gabriel Synnaeve, Nicolas Usunier, Alexander
  Kirillov, and Sergey Zagoruyko.
\newblock End-to-end object detection with transformers.
\newblock In {\em European Conference on Computer Vision (ECCV)}, 2020.

\bibitem{8126154}
N. Deepika and V.~V. Sajith~Variyar.
\newblock Obstacle classification and detection for vision based navigation for
  autonomous driving.
\newblock In {\em International Conference on Advances in Computing,
  Communications and Informatics (ICACCI)}, 2017.

\bibitem{denouden2018improving}
Taylor Denouden, Rick Salay, Krzysztof Czarnecki, Vahdat Abdelzad, Buu Phan,
  and Sachin Vernekar.
\newblock Improving reconstruction autoencoder out-of-distribution detection
  with mahalanobis distance.
\newblock {\em ArXiv}, abs/1812.02765, 2018.

\bibitem{dhamija2020overlooked}
Akshay Dhamija, Manuel Gunther, Jonathan Ventura, and Terrance Boult.
\newblock The overlooked elephant of object detection: Open set.
\newblock In {\em IEEE/CVF Winter Conference on Applications of Computer Vision
  (WACV)}, 2020.

\bibitem{dosovitskiy2017carla}
Alexey Dosovitskiy, German Ros, Felipe Codevilla, Antonio Lopez, and Vladlen
  Koltun.
\newblock Carla: An open urban driving simulator.
\newblock In {\em Conference on robot learning (CORL)}, 2017.

\bibitem{du2022unknown}
Xuefeng Du, Xin Wang, Gabriel Gozum, and Yixuan Li.
\newblock Unknown-aware object detection: Learning what you don't know from
  videos in the wild.
\newblock In {\em IEEE/CVF Conference on Computer Vision and Pattern
  Recognition (CVPR)}, 2022.

\bibitem{vos}
Xuefeng Du, Zhaoning Wang, Mu Cai, and Yixuan Li.
\newblock Vos: Learning what you don’t know by virtual outlier synthesis.
\newblock In {\em International Conference on Learning Representations (ICLR)},
  2022.

\bibitem{voc}
Mark Everingham, Luc Van~Gool, Christopher~KI Williams, John Winn, and Andrew
  Zisserman.
\newblock The pascal visual object classes (voc) challenge.
\newblock {\em International Journal of Computer Vision (IJCV)},
  88(2):303--338, 2010.

\bibitem{gal2016dropout}
Yarin Gal and Zoubin Ghahramani.
\newblock Dropout as a bayesian approximation: Representing model uncertainty
  in deep learning.
\newblock In {\em International Conference on Machine Learning (ICML)}, 2016.

\bibitem{OWDETR}
Akshita Gupta, Sanath Narayan, KJ Joseph, Salman Khan, Fahad~Shahbaz Khan, and
  Mubarak Shah.
\newblock {OW-DETR}: Open-world detection transformer.
\newblock In {\em IEEE/CVF Conference on Computer Vision and Pattern
  Recognition (CVPR)}, 2022.

\bibitem{hall2020probabilistic}
David Hall, Feras Dayoub, John Skinner, Haoyang Zhang, Dimity Miller, Peter
  Corke, Gustavo Carneiro, Anelia Angelova, and Niko S{\"u}nderhauf.
\newblock Probabilistic object detection: Definition and evaluation.
\newblock In {\em IEEE/CVF Winter Conference on Applications of Computer Vision
  (WACV)}, 2020.

\bibitem{he2017mask}
Kaiming He, Georgia Gkioxari, Piotr Doll{\'a}r, and Ross Girshick.
\newblock Mask r-cnn.
\newblock In {\em IEEE International Conference on Computer Vision (ICCV)},
  2017.

\bibitem{he2016deep}
Kaiming He, Xiangyu Zhang, Shaoqing Ren, and Jian Sun.
\newblock Deep residual learning for image recognition.
\newblock In {\em IEEE Conference on Computer Vision and Pattern Recognition
  (CVPR)}, 2016.

\bibitem{hendrycks2016baseline}
Dan Hendrycks and Kevin Gimpel.
\newblock A baseline for detecting misclassified and out-of-distribution
  examples in neural networks.
\newblock {\em ArXiv}, abs/1610.02136, 2016.

\bibitem{janai2020computer}
Joel Janai, Fatma G{\"u}ney, Aseem Behl, Andreas Geiger, et~al.
\newblock Computer vision for autonomous vehicles: Problems, datasets and state
  of the art.
\newblock {\em Foundations and Trends{\textregistered} in Computer Graphics and
  Vision (FTCGV)}, 12(1--3):1--308, 2020.

\bibitem{owod}
KJ Joseph, Salman Khan, Fahad~Shahbaz Khan, and Vineeth~N Balasubramanian.
\newblock Towards open world object detection.
\newblock In {\em IEEE/CVF Conference on Computer Vision and Pattern
  Recognition (CVPR)}, 2021.

\bibitem{liang2017enhancing}
Shiyu Liang, Yixuan Li, and Rayadurgam Srikant.
\newblock Enhancing the reliability of out-of-distribution image detection in
  neural networks.
\newblock {\em ArXiv}, abs/1706.02690, 2017.

\bibitem{lin2017feature}
Tsung-Yi Lin, Piotr Doll{\'a}r, Ross Girshick, Kaiming He, Bharath Hariharan,
  and Serge Belongie.
\newblock Feature pyramid networks for object detection.
\newblock In {\em IEEE Conference on Computer Vision and Pattern Recognition
  (CVPR)}, 2017.

\bibitem{lin2017focal}
Tsung-Yi Lin, Priya Goyal, Ross Girshick, Kaiming He, and Piotr Doll{\'a}r.
\newblock Focal loss for dense object detection.
\newblock In {\em IEEE International Conference on Computer Vision (ICCV)},
  2017.

\bibitem{lin2014microsoft}
Tsung-Yi Lin, Michael Maire, Serge Belongie, James Hays, Pietro Perona, Deva
  Ramanan, Piotr Doll{\'a}r, and C~Lawrence Zitnick.
\newblock {Microsoft COCO}: Common objects in context.
\newblock In {\em European Conference on Computer Vision (ECCV)}, 2014.

\bibitem{liu2016ssd}
Wei Liu, Dragomir Anguelov, Dumitru Erhan, Christian Szegedy, Scott Reed,
  Cheng-Yang Fu, and Alexander~C Berg.
\newblock Ssd: Single shot multibox detector.
\newblock In {\em European Conference on Computer Vision (ECCV)}, 2016.

\bibitem{liu2020energy}
Weitang Liu, Xiaoyun Wang, John Owens, and Yixuan Li.
\newblock Energy-based out-of-distribution detection.
\newblock {\em Advances in Neural Information Processing Systems (NIPS)}, 2020.

\bibitem{ZheLiu-AAAI2019-TANet}
Zhe Liu, Xin Zhao, Tengteng Huang, Ruolan Hu, Yu Zhou, and Xiang Bai.
\newblock Tanet: Robust 3d object detection from point clouds with triple
  attention.
\newblock In {\em AAAI Conference on Artificial Intelligence(AAAI)}, 2020.

\bibitem{OLP}
Jianxiang Ma, Anlong Ming, Zilong Huang, Xinggang Wang, and Yu Zhou.
\newblock Object-level proposals.
\newblock In {\em IEEE International Conference on Computer Vision (ICCV)},
  2017.

\bibitem{miller2019evaluating}
Dimity Miller, Feras Dayoub, Michael Milford, and Niko S{\"u}nderhauf.
\newblock Evaluating merging strategies for sampling-based uncertainty
  techniques in object detection.
\newblock In {\em International Conference on Robotics and Automation (ICRA)},
  2019.

\bibitem{miller2018dropout}
Dimity Miller, Lachlan Nicholson, Feras Dayoub, and Niko S{\"u}nderhauf.
\newblock Dropout sampling for robust object detection in open-set conditions.
\newblock In {\em IEEE International Conference on Robotics and Automation
  (ICRA)}, 2018.

\bibitem{pinggera2016lost}
Peter Pinggera, Sebastian Ramos, Stefan Gehrig, Uwe Franke, Carsten Rother, and
  Rudolf Mester.
\newblock Lost and found: detecting small road hazards for self-driving
  vehicles.
\newblock In {\em IEEE/RSJ International Conference on Intelligent Robots and
  Systems (IROS)}, 2016.

\bibitem{ramos2017detecting}
Sebastian Ramos, Stefan Gehrig, Peter Pinggera, Uwe Franke, and Carsten Rother.
\newblock Detecting unexpected obstacles for self-driving cars: Fusing deep
  learning and geometric modeling.
\newblock In {\em IEEE Intelligent Vehicles Symposium (IV)}, 2017.

\bibitem{redmon2016you}
Joseph Redmon, Santosh Divvala, Ross Girshick, and Ali Farhadi.
\newblock You only look once: Unified, real-time object detection.
\newblock In {\em IEEE Conference on Computer Vision and Pattern Recognition
  (CVPR)}, 2016.

\bibitem{redmon2017yolo9000}
Joseph Redmon and Ali Farhadi.
\newblock Yolo9000: better, faster, stronger.
\newblock In {\em IEEE Conference on Computer Vision and Pattern Recognition
  (CVPR)}, 2017.

\bibitem{ren2015faster}
Shaoqing Ren, Kaiming He, Ross Girshick, and Jian Sun.
\newblock Faster r-cnn: Towards real-time object detection with region proposal
  networks.
\newblock In {\em Advances in Neural Information Processing Systems (NIPS)},
  2015.

\bibitem{Ribeiro2014}
Henrique~Jales Ribeiro.
\newblock {\em The Role of Analogy in Philosophical Discourse}.
\newblock Springer International Publishing, 2014.

\bibitem{ncut}
Jianbo Shi and Jitendra Malik.
\newblock Normalized cuts and image segmentation.
\newblock {\em IEEE Transactions on Pattern Analysis and Machine Intelligence
  (TPAMI)}, 22(8):888--905, 2000.

\bibitem{wu2019detectron2}
Yuxin Wu, Alexander Kirillov, Francisco Massa, Wan-Yen Lo, and Ross Girshick.
\newblock Detectron2.
\newblock \url{https://github.com/facebookresearch/detectron2}, 2019.

\bibitem{wu2022uc}
Zhiheng Wu, Yue Lu, Xingyu Chen, Zhengxing Wu, Liwen Kang, and Junzhi Yu.
\newblock {UC-OWOD}: Unknown-classified open world object detection.
\newblock In {\em European Conference on Computer Vision (ECCV)}, 2022.

\bibitem{8794279}
Feng Xue, Anlong Ming, Menghan Zhou, and Yu Zhou.
\newblock A novel multi-layer framework for tiny obstacle discovery.
\newblock In {\em International Conference on Robotics and Automation (ICRA)},
  2019.

\bibitem{9210191}
Feng Xue, Anlong Ming, and Yu Zhou.
\newblock Tiny obstacle discovery by occlusion-aware multilayer regression.
\newblock {\em IEEE Transactions on Image Processing (TIP)}, 29:9373--9386,
  2020.

\bibitem{yang2021objects}
Shuo Yang, Peize Sun, Yi Jiang, Xiaobo Xia, Ruiheng Zhang, Zehuan Yuan, Changhu
  Wang, Ping Luo, and Min Xu.
\newblock Objects in semantic topology.
\newblock In {\em International Conference on Learning Representations (ICLR)},
  2022.

\bibitem{yu2020bdd100k}
Fisher Yu, Haofeng Chen, Xin Wang, Wenqi Xian, Yingying Chen, Fangchen Liu,
  Vashisht Madhavan, and Trevor Darrell.
\newblock {Bdd100k}: A diverse driving dataset for heterogeneous multitask
  learning.
\newblock In {\em IEEE/CVF Conference on Computer Vision and Pattern
  Recognition (CVPR)}, 2020.

\bibitem{rowod}
Xiaowei Zhao, Xianglong Liu, Yifan Shen, Yixuan Qiao, Yuqing Ma, and Duorui
  Wang.
\newblock Revisiting open world object detection.
\newblock {\em ArXiv}, abs/2201.00471, 2022.

\bibitem{NIPS2012_3e313b9b}
Yu Zhou, Xiang Bai, Wenyu Liu, and Longin Latecki.
\newblock Fusion with diffusion for robust visual tracking.
\newblock In {\em Advances in Neural Information Processing Systems (NIPS)},
  2012.

\bibitem{YuZhou-IJCV2016-SFVT}
Yu Zhou, Xiang Bai, Wenyu Liu, and Longin~Jan Latecki.
\newblock Similarity fusion for visual tracking.
\newblock {\em International Journal of Computer Vision (IJCV)},
  118(3):337--363, 2016.

\bibitem{2014ONLINE}
Yu Zhou., Yinfei Yang., Yi Meng., Xiang Bai, Wenyu Liu, and Longin~Jan Latecki.
\newblock Online multiple person detection and tracking from mobile robot in
  cluttered indoor environments with depth camera.
\newblock {\em International Journal of Pattern Recognition and Artificial
  Intelligence (IJPRAI)}, 28(1):1455001.1--1455001.28, 2014.

\bibitem{zhu2020deformable}
Xizhou Zhu, Weijie Su, Lewei Lu, Bin Li, Xiaogang Wang, and Jifeng Dai.
\newblock Deformable detr: Deformable transformers for end-to-end object
  detection.
\newblock In {\em International Conference on Learning Representations (ICLR)},
  2021.

\bibitem{edgeboxes}
C.~Lawrence Zitnick and Piotr Doll{\'a}r.
\newblock Edge boxes: Locating object proposals from edges.
\newblock In David Fleet, Tomas Pajdla, Bernt Schiele, and Tinne Tuytelaars,
  editors, {\em European Conference on Computer Vision (ECCV)}, 2014.

\end{thebibliography}
}

\end{document}